\documentclass[10pt,twocolumn,letterpaper]{article}

\usepackage{wacv}
\usepackage{times}
\usepackage{epsfig}
\usepackage{graphicx}
\usepackage[tbtags]{amsmath}
\usepackage{amssymb}
\usepackage{booktabs}
\usepackage[T1]{fontenc}
\usepackage[dvipsnames, table]{xcolor}
\usepackage{dirtytalk}

\usepackage{bm}
\usepackage{physics}
\usepackage{nicefrac}
\usepackage{cite}

\newtheorem{remark}{Remark}

\usepackage{multirow}

\usepackage{algorithm}
\usepackage{algpseudocodex}

\usepackage{tikz}
\usetikzlibrary{arrows.meta}

\usepackage[normalem]{ulem}
\usepackage{tablefootnote}
\usepackage{authblk}
\def\wacvPaperID{19} % Enter the WACV Paper ID here

\wacvalgorithmstrack   % Uncomment this line if you are submitting to the Algorithms Track.
\wacvfinalcopy % *** Uncomment this line for the final submission

\usepackage[pagebackref=true, breaklinks=true, colorlinks, citecolor=PineGreen, linkcolor=BrickRed, bookmarks=false]{hyperref}
\usepackage[capitalize]{cleveref}

\begin{document}

    %%%%%%%%% TITLE
    \title{Instance-Dependent Noisy Label Learning via Graphical Modelling}
    
    % \author{Arpit Garg \\
    % Australian Institute for Machine Learning \\
    % University of Adelaide, Australia \\
    % {\tt\small arpit.garg@adelaide.edu.au}
    % \and
    \author[1]{Arpit Garg\thanks{arpit.garg@aiml.team}}
    \author[1]{Cuong Nguyen}
    \author[1]{Rafael Felix}
    \author[2]{Thanh-Toan Do}
    \author[1]{Gustavo Carneiro}
    \affil[1]{Australian Institute for Machine Learning, University of Adelaide}
    \affil[2]{Department of Data Science and AI, Faculty of Information Technology, Monash University}
    
    \renewcommand\Affilfont{\normalsize}
    
    \maketitle
    \thispagestyle{empty}
    
    %%%%%%%%% ABSTRACT
    \begin{abstract}
    Noisy labels are unavoidable yet troublesome in the ecosystem of deep learning because models can easily overfit them. There are many types of label noise, such as symmetric, asymmetric and instance-dependent noise (IDN), with IDN being the only type that depends on image information. Such dependence on image information makes IDN a critical type of label noise to study, given that labelling mistakes are caused in large part by insufficient or ambiguous information about the visual classes present in images. Aiming to provide an effective technique to address IDN, we present a new graphical modelling approach called InstanceGM, that combines  discriminative and generative models. The main contributions of InstanceGM are: i) the use of the continuous Bernoulli distribution to train the generative model, offering significant training advantages, and ii) the exploration of a state-of-the-art noisy-label discriminative classifier to generate clean labels from instance-dependent noisy-label samples. InstanceGM is competitive with current noisy-label learning approaches, particularly in IDN benchmarks using synthetic and real-world datasets, where our method shows better accuracy than the competitors in most experiments.
\end{abstract}

    %%%%%%%%% BODY TEXT
    \section{Introduction}

    The latest developments in deep neural networks (DNNs) have shown outstanding results in a variety of applications ranging from computer vision~\cite{krizhevsky2012imagenet} to natural language processing~\cite{ragesh2021hetegcn} and medical image analysis~\cite{qian20223d}. Such success is strongly reliant on high-capacity models, which in turn, require a massive amount of correctly-annotated data for training~\cite{yao2020dual, litjens2017survey}. Annotating a large amount of data is, however, arduous, costly and time-consuming, and therefore is often done via crowd-sourcing~\cite{song2022learning} that generally produces low-quality annotations. Although that brings down the cost and scales up the process, the trade-off is the mislabelling of the data, resulting in a deterioration of deep models' performance~\cite{liu2020early, bae2022noisy} due to the \emph{memorisation effect}~\cite{neyshabur2017exploring, arpit2017closer, liu2020early, zhang2021learning_}. This has, therefore, motivated the research of novel learning algorithms to tackle the label noise problem where data might have been mislabelled.
    
    % The degradation in the performance of various deep learning applications are due to the presence of noisy labels~\cite{bae2022noisy, liu2020early} and it could be due to labelling process including human mistakes~\cite{cheng2020learning}, crowd-sourcing~\cite{song2022learning}, low-quality data~\cite{frenay2013classification}. Due to the over-fitting power of DNNs, the incorrect labels for the instances can be easily memorised also known as \emph{memorization effect}~\cite{liu2020early, zhang2021learning_,neyshabur2017exploring,arpit2017closer}. Label noise robust algorithms are very crucial for real-world applications, where incorrect annotations are very common in training~\cite{cordeiro2021propmix}. In Class-conditional noise (CCN), each class instance is assigned with a set probability whereas in Instance-Dependent noise (IDN), mislabeling depends on an instance's class and characteristics. This work focuses on IDN, a generic noise and real-world noise method.

    %The label noise problem has been studied from 1990s~\cite{angluin1988learning} and recently attracted much of research interest. 
    Early work in label noise~\cite{han2018co} was carried out under the assumption that label noise was instance-independent (IIN), i.e., mislabelling occurred regardless of the information about the visual classes present in images.
    In IIN, we generally have a transition matrix that contains a pre-defined probability of flipping between pairs of labels (e.g., any image showing a \emph{cat} has a high priori probability of being mislabelled as a \emph{dog} and low a priori probability of being mislabelled as a \emph{car}).
    %e.g. \(p(\hat{Y} | X, Y) = p(\hat{Y} | Y)\) where \(\hat{Y}, Y\) and \(X\) are random variables denoting the noisy label, clean label and the image, respectively. 
    This type of noise can also be divided into two sub-types: \emph{symmetric}, where a true label is flipped to another label with equal probability across all classes, and \emph{asymmetric}, where a true label is more likely to be mislabeled into one of some particular classes~\cite{han2018co}. 
    Nevertheless, the IIN assumption is impractical for many real-world datasets because we can intuitively argue that mislabellings mostly occur because of insufficient or ambiguous information  about the visual classes present in images. 
    As a result, recent studies have gradually shifted their focus toward the more realistic scenario of instance-dependent noise (IDN), where label noise depends on both the true class label and the image information~\cite{xia2020part}.

    Many methods have been introduced to handle not only IIN, but also IDN problems. Those include, but are not limited to, \emph{sample selection}~\cite{xia2021sample, li2020dividemix, zheltonozhskii2022contrast,kim2021fine, cordeiro2021propmix} that detects clean and noisy labels and applies semi-supervised learning methods on the processed data, \emph{robust losses}~\cite{patrini2017making, arazo2019unsupervised, liu2020peer} that can work well with either clean or noisy labels, and \emph{probabilistic approaches}~\cite{yao2021instance} that model the data generation process, including how a noisy label is created. 
    Despite some successes, most methods are often demonstrated in IIN settings with simulated symmetric and asymmetric noise. However, their performance is degraded when evaluated on IDN problems, which include real-world and synthetic datasets. Although there are a few studies focusing on the IDN setting~\cite{yao2021instance, cheng2022instance, xia2020part,zhu2021clusterability, jiang2021information}, their relatively inaccurate classification results suggest that the algorithms can be improved further.

In this paper, we propose a new method to tackle the IDN problem, called InstanceGM. Our method is designed based on a graphical model that considers the clean label \(Y\) as a latent variable and introduces another latent variable \(Z\) representing the image feature to model the generation of a label noise \(\hat{Y}\) and an image \(X\). 
InstanceGM integrates generative and discriminate models, where the generative model is based on a variational auto-encoder (VAE)~\cite{kingma2013auto}, except that we replace the conventional mean squared error (MSE) when modelling the likelihood of reconstructed images by a  \emph{continuous Bernoulli distribution}~\cite{loaiza2019continuous} that facilitates the training process since it avoids tuning additional hyper-parameters. 
%Furthermore, instead of relying on co-teach~\cite{han2018co} as in CausalNL~\cite{yao2021instance}, that has a severe limitation on selecting samples excluding the noisily-labelled data, we integrate DivideMix~\cite{li2020dividemix} into our learning algorithm. 
For the discriminative model, to mitigate the problem of 
only using clean label data during the training process, which is a common issue present in the similar graphical model methods~\cite{yao2021instance}, we rely on  DivideMix~\cite{li2020dividemix} that uses both clean and noisy-label data for training by exploring semi-supervised learning via MixMatch~\cite{berthelot2019mixmatch}. DivideMix is shown to be a reasonably effective discriminative classifier for our InstanceGM.
%As DivideMix is designed to not only filters out clean and noisy labels, but also uses semi-supervised learning via MixMatch~\cite{berthelot2019mixmatch} to utilise the availability of un-labelled data obtained from the noisy labels, it helps to improve further the performance, especially on the instance-dependent noise setting. 
In summary, the main contributions of the proposed method are:
% In this paper, we propose a new instance-dependent noisy label learning algorithm that addresses the before-mentioned points. It filters out the clean and noisy samples and use DivideMix~\cite{li2020dividemix}, and then considered the same set of classifiers for modeling the clean labels and new set of encoders for learning latent representation. Then, used two different set of decoders for the reconstruction of the image instances and prediction of the noisy labels using the continuous Bernoulli loss and classification loss respectively. Finally, we conduct various experiments on various datasets to elucidate that proposed methos is superior to previous approached for IDN robust learning.
% The main contributions of the proposed work are summarised as follows:
\begin{itemize}
    \item InstanceGM follows a graphical modelling approach to generate both the image \(X\) and its noisy label \(\hat{Y}\) with the true label \(Y\) and image feature \(Z\) as latent variables. The modelling is associated with the continuous Bernoulli distribution to model the generation of instance \(X\) to facilitate the training, avoiding tuning of additional hyper-parameters (see \cref{remark:avoid_reweighting}).
    % We implemented generative modelling using the set of encoders and decoders. We have two networks which trains simultaneously and consists of classifiers to generate the clean labels and set of encoders to generate the latent representations. Additionally, the reconstruction of instances are done utilizing the decoders. 
    \item For the discriminative classifier of InstanceGM, we replace the commonly used
    co-teaching, which is a dual model that relies only on training samples classified as clean, with DivideMix~\cite{li2020dividemix} that uses all training samples classified as clean and noisy. 
    %instead of selecting a few that have a high chance of being clean.
    % In the classifier stage we use DivideMix and improve the reconstruction pervasive error using continuous Bernoulli distribution. Moreover, we provide the ablation study to examine the effect of various components.
    \item InstanceGM shows state-of-the-art results on a variety of IDN benchmarks, including simulated and real-world datasets, such as CIFAR10 and CIFAR100~\cite{krizhevsky2009learning}, Red Mini-ImageNet from Controlled Noisy Web Labels (CNWL)~\cite{xu2021faster}, ANIMAL-10N~\cite{song2019selfie} and CLOTHING-1M~\cite{xiao2015learning}.
    % \item Empirical results on CIFAR10 and CIFAR100~\cite{krizhevsky2009learning}  benchmark dataset under instance-dependent noise, showcase that our proposed algorithm outperforms previous approaches. For high-noise rate (IDN) and real-world noise problems in CIFAR10/CIFAR100~\cite{krizhevsky2009learning}, Red Mini-ImageNet from CNWL~\cite{xu2021faster}, ANIMAL-10N~\cite{song2019selfie} and CLOTHING-1M~\cite{xiao2015learning}, our approach presents among the best results in the field by a notable margin. 
\end{itemize}

    \section{Related work}
\label{sec:related_work}

As DNNs have been shown to easily fit randomly labelled training data~\cite{zhang2017understanding}, they can also overfit a noisy-label dataset, which eventually results in poor generalisation to a clean-label testing data~\cite{neyshabur2017exploring, arpit2017closer, liu2020early, zhang2021learning_}. 
Several studies have, therefore, been conducted to investigate supervised learning under the label noise setting, including robust loss function~\cite{ma2020normalized, wang2019imae}, sample selection~\cite{wang2018iterative, song2019selfie, song2021robust}, robust regularisation~\cite{jenni2018deep, wei2021open, menon2019can, goodfellow2014explaining} and robust architecture~\cite{xiao2015learning, cheng2020weakly, han2018masking, kong2021resolving}. Below, we review methods dealing with noisy labels, especially IDN, without the reliance on clean validation sets~\cite{veit2017learning, hendrycks2018using, ren2018learning}.

%There have been multiple papers written about the label memorisation ability of deep neural networks and the relationship between memorisation and generalisation in terms of noisy labels~\cite{liu2020early, zhang2021learning_,neyshabur2017exploring,arpit2017closer}. Various studies proposed methods to address such label memorisation issues~\cite{song2019does, han2020sigua,cheng2020learning,yao2020searching}. Nevertheless, there is still a poor understanding in the field of why one strategy works or when fails, despite the fact that several approaches seem to be successful in addressing various types of noise~\cite{liu2021understanding}.

Let us start with methods designed to handle \say{any} type of label noise, including IDN and IIN.
An important technique for both of these label noise types is sample selection~\cite{wang2018iterative, song2019selfie, song2021robust}, which aims to select clean-label samples automatically for training. Although it is well-motivated and often effective, it suffers from the cumulative error caused by mistakes in the selection process, mainly when there are numerous unclear classes in the training data. Consequently, sample selection methods often rely on multiple clean-label sample classifiers to increase their robustness against such cumulative error~\cite{li2020dividemix}. 
In addition, semi-supervised learning (SSL)~\cite{cordeiro2021propmix, li2020dividemix, song2019selfie, zheltonozhskii2022contrast, kim2021fine, bernhardt2022active} have also been integrated with sample selection and multiple clean-label classifiers to enable the training from clean and noisy-label samples. 
In particular, SSL methods use clean and noisy samples by treating them as labelled and unlabelled data, respectively, with a MixMatch approach~\cite{berthelot2019mixmatch}. These methods above have been designed to handle \say{any} type of label noise, so they are usually assessed in synthetic IIN benchmarks and real-world IDN benchmarks.

Given that real-world datasets do not, in general, contain IIN, more recently proposed methods aim to address IDN benchmarks~\cite{cheng2022instance, xia2020part, Zhu_2021_CVPR, berthon2021confidence, yao2021instance, liu2021understanding}. 
In these benchmarks, the task of differentiating between hard clean-labelled samples and noisy-label samples pose a major challenge.
Such issue is noted by Song et al.~\cite{song2019does}, who state that the model performance in IDN can degrade significantly compared to other types of noises.

One direct way of addressing IDN problems relies on a graphical model approach that has random variables representing the observed noisy label, the image, and the latent clean label.
This model also has a generative process to produce an image given the (clean and noisy) label information~\cite{lawrence2001estimating}. Another approach examines a graphical model using a discriminative process~\cite{raykar2010learning},  where the model attempts to explain the posterior probability of the observed noisy label by averaging the posterior probabilities of the clean class label. 
% Using causal information to contribute to the selection of clean labels, 
Yao et al.~\cite{yao2021instance} developed a new causal model to address IDN that also uses the same variables as the methods above plus a latent image feature variable, which relies on  generative models to produce the image from the clean label and image feature, and to produce the noisy label from the image feature and clean label.
That approach~\cite{yao2021instance}, however, did not produce competitive results compared with state of the art.
We argue that the model's poor performance is mostly due to the co-teaching~\cite{han2018co} that is trained with a small set of samples classified as clean, which can inadvertently contain noisy-label samples -- this is an issue that can cause a cumulative error, particularly in IDN problems.
%was heavily reliant on. In addition, by using co-teaching, our model included knowledge from a tiny batch of data while ignoring the rest.

Our work is motivated by the graphical model approaches mentioned above, that aim to address IDN problems. 
%and stresses comprehension at the instance level. This emphasis is well suited for a research with lengthy ranges of occurrences that occur at varying rates, as is often shown with visual datasets.
The main difference in our approach is the use of a more effective clean sample identifier that replaces  co-teaching~\cite{yao2021instance} by DivideMix~\cite{li2020dividemix}, which considers the whole training set, instead of only the samples classified as clean.
Moreover, we propose a more effective training of the image generative model based on the continuous Bernoulli distribution~\cite{loaiza2019continuous}. 

%reconstruction loss, as due to the fact that VAE tries to regress the pixel intensity. We show in the experiments that our model achieves great result with high accuracy. 

    \section{Methodology}

\label{sec:methodology}

    \begin{figure}[t]
        \centering
        \begin{tikzpicture}
            \input{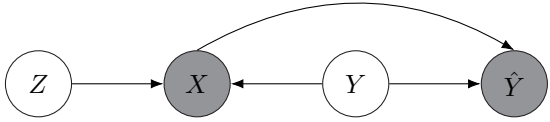}
        \end{tikzpicture}
        \caption{The proposed graphical model of the generation process that produces the observable (shaded nodes) data \(X\) and noisy label \(\hat{Y}\) from hidden (non-shaded nodes) data representation \(Z\) and clean label \(Y\).}
        \label{fig:graphical_model}
    \end{figure}

    \subsection{Problem definition}
    \label{sec:background}
        We denote \(X\)
        as an observed random variable representing an image, 
        \(Y\)
        as a latent random variable corresponding to the clean label of \(X\),
        \(Z\) as a latent random variable denoting an image feature representation for of \(X\), and \(\hat{Y}\) as the observed random variable for the noisy label.
        The training set is represented by \(\mathcal{D}=\{(\mathbf{x}_i,\hat{\mathbf{y}}_i)\}_{i=1}^{|\mathcal{D}|}\), where 
        the image is represented by \(\mathbf{x} \in \mathcal{X} \subset \mathbb{R}^{H \times W \times 3}\) (with \(3\) color channels and size \(H \times W\) pixels) and the noisy label $\hat{\mathbf{y}} \in \mathcal{Y} \in \{0,1\}^{|\mathcal{Y}|}$ denoted by a one-hot vector.
        %In image classification, the input instance \(X\) denotes an RGB image of size \(H \times W \times 3\) and the corresponding label \(Y\) is an one-hot vector. 
        In the conventional supervised learning, \(\mathcal{D}\) is used to train a model \(f_{\theta}: \mathcal{X} \to \Delta^{|\mathcal{Y}|-1}\) (where \(\Delta^{|\mathcal{Y}|-1}\) represents the probability simplex), parameterised by \(\theta \in \Theta\), that can predict the labels of testing images. 
        %In the label noise setting, the clean label \(Y\) is assumed to be hidden, and a noisy one -- denoting as \(\hat{Y}\) -- is observed. 
        The aim is to exploit the noisy data \((X, \hat{Y})\) from a training set to infer a model \(f_{\theta}\) that can accurately predict the clean labels \(Y\) of data in a testing set.

    \subsection{Probabilistic noisy label modelling}
    \label{sec:probabilistic_modelling}
        We follow a similar approach presented in~\cite{yao2021instance} to model the process that generates samples with noisy labels via the graphical model shown in \cref{fig:graphical_model}, where the clean label \(Y\) and image feature representation \(Z\) are latent variables. Under this modelling assumption, a noisy-label sample \((\mathbf{x}, \hat{\mathbf{y}})\) can be generated as follows:
        \begin{enumerate}
            \item sample a clean label from its prior: \(\mathbf{y} \sim p(Y)\),
            \item sample a representation from its prior: \(\mathbf{z} \sim p(Z)\),
            \item sample an input data from its continuous Bernoulli distribution: \(\hat{\mathbf{x}} \sim \mathcal{CB}(X; \lambda(\mathbf{z}, \mathbf{y}))\), \label{forward:generate_x}
            \item sample the corresponding noisy label from its categorical distribution: \(\hat{\mathbf{y}} \sim \mathrm{Cat}(\hat{Y}; \gamma(\hat{\mathbf{x}}, \mathbf{y}))\)
            % \gustavo{According to Figure~\ref{fig:graphical_model}, we have a discrepancy here because $\hat{Y}$ depends on $X$ and $Y$, not $Z$ and $Y$. \(\hat{\mathbf{y}} \sim \mathrm{Cat}(\hat{Y}; \gamma(\hat{\mathbf{x}}, \mathbf{y}))\)}.
        \end{enumerate}

        \begin{remark}\label{remark}
            Conventionally, the process of generating data \(\mathbf{x}\) in step~\ref{forward:generate_x} above is often modelled as a Bernoulli distribution or multivariate normal distribution, corresponding to the binary cross-entropy (BCE) or MSE reconstruction losses, respectively. Such modelling, however, leads to a pervasive error~\cite{loaiza2019continuous} since the image pixels are in \([0, 1]\) instead of \(\{0, 1\}\) (Bernoulli distribution)\footnote{Except for black and white images.} or \((-\infty, +\infty)\) (multivariate normal distribution). We therefore adopt the continuous Bernoulli distribution~\cite{loaiza2019continuous} which has a support in \([0, 1]\) to correctly model this image generation process.
        \end{remark}
        
        Note that the parameters of the continuous Bernoulli and categorical distributions are conditioned on \(Z\), \(X\) and \(Y\), and modelled as the outputs of two DNNs:
        \begin{equation}
            \lambda = f_{\theta_{x}}(\mathbf{z, y}) \quad \mathrm{and} \quad \gamma = f_{\theta_{\hat{y}}} (\hat{\mathbf{x}}, \mathbf{y}),
            \label{eq:probs_decoders}
        \end{equation}
        % \gustavo{similarly, shouldn't we have $\gamma = f_{\theta_{\hat{y}}} (\hat{\mathbf{x}}, \mathbf{y})$.}
        where \(f\) denotes the neural network,
        and \(\theta_{x}\),\(\theta_{\hat{y}}\) represent the network parameters. Following the convention in machine learning, we call \(f_{\theta_{x}}(.)\) the \emph{decoder} and \(f_{\theta_{\hat{y}}}(.)\) the \emph{noisy label classifier}.

        To solve the label noise problem that has data generated from the process above, we need to infer the posterior \(p(Z, Y | X, \hat{Y})\). However, due to the complexity of the graphical model in \cref{fig:graphical_model}, exact inference for the posterior \(p(Z, Y | X, \hat{Y})\) is intractable, and therefore, the estimation must rely on an approximation. Motivated by~\cite{yao2021instance}, we employ variational inference to approximate the true posterior \(p(Z, Y | X, \hat{Y})\) by a variational \say{posterior} \(q(Z, Y | X, \hat{Y})\). Such posterior can be obtained by minimising the following Kullback-Leibler (KL) divergence:
        \begin{equation}
            \begin{aligned}[b]
                & \min_{q} \mathrm{KL} \left[ q(Z, Y | X, \hat{Y}) \, || \, p(Z, Y | X, \hat{Y}) \right],%\\
                % & = \min_{q} \mathbb{E}_{q(Z, Y | X, \hat{Y})} \left[ - \ln p(X | Z, Y) - \ln p(\hat{Y} | Z, Y) \right] + \mathrm{KL} \left[ q(Z, Y | X, \hat{Y}) \, || \, p(Z, Y) \right].
            \end{aligned}
            \label{eq:minimise_KL_divergence}
        \end{equation}
        where the variational posterior \(q(Z, Y | X, \hat{Y})\) can be factorised following the product rule of probability. We assume that the posterior of the clean label \(Y\) is independent from the noisy label \(\hat{Y}\), given the instance \(X\): \(q(Y | X, \hat{Y}) = q(Y | X)\). In addition, the variational posterior of feature representation is independent from the noisy label given the clean label and input data: \(q(Z | X, \hat{Y}, Y)  = q(Z | X, Y)\). The variational posterior of interest can, therefore, be written as:
        \begin{equation}
            \begin{split}
               q(Z, Y | X, \hat{Y}) &= q(Z | X, \hat{Y}, Y) \, q(Y | X, \hat{Y}) \\ 
               & = q(Z | X, Y) \, q(Y | X).  \label{eq:posterior}
            \end{split}
        \end{equation}

        \begin{figure*}[ht!]
            \centering
            \includegraphics[width=0.65\linewidth, keepaspectratio]{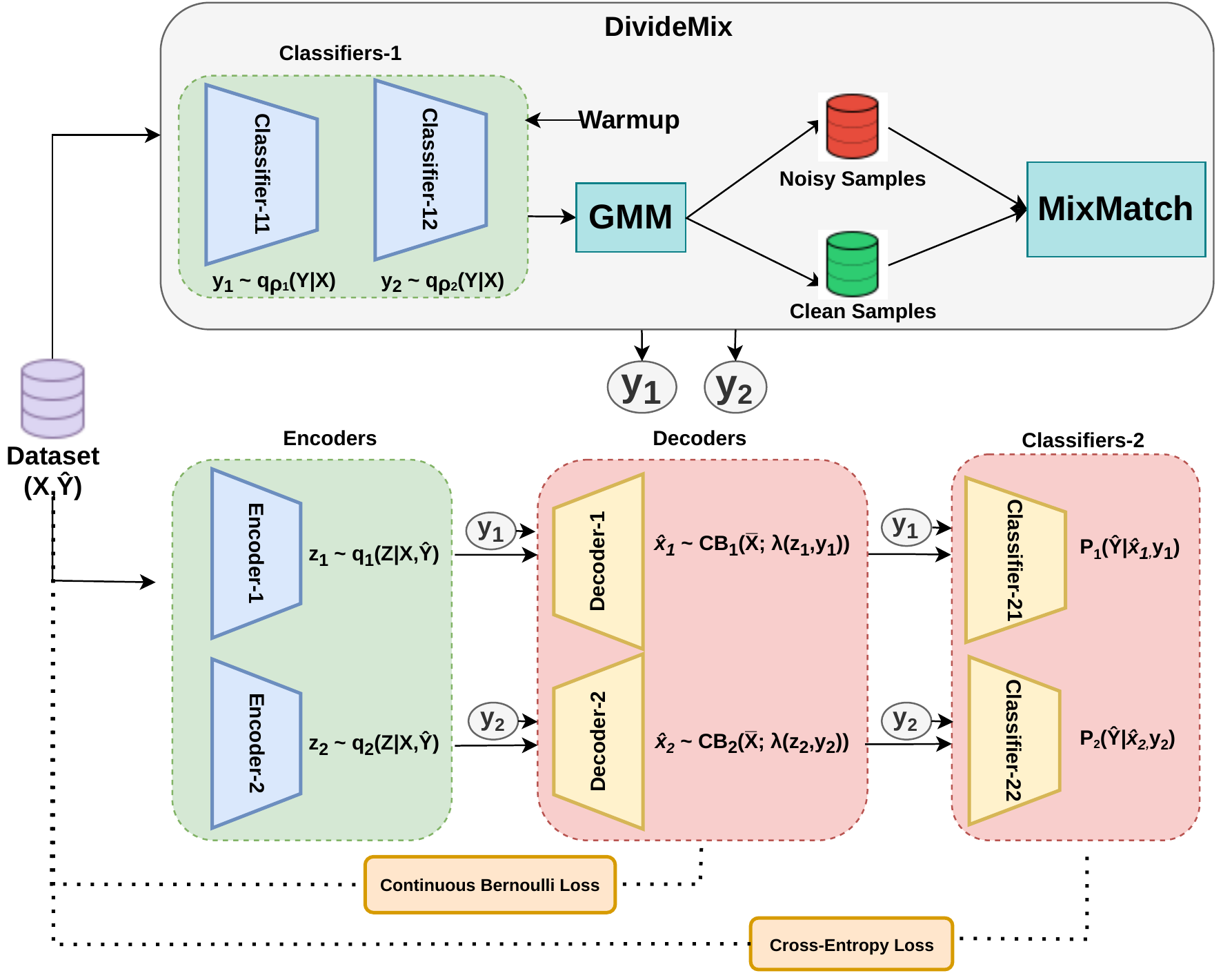}
            \caption{The proposed InstanceGM trains the \textit{Classifiers} to output clean labels for  instance-dependent noisy-label samples. We first warmup our two classifiers \textit{(Classifier-\{11,12\})} using the classification loss, and then with classification loss we train the GMM to separate clean and noisy samples with the semi-supervised model MixMatch~\cite{berthelot2019mixmatch} from the DivideMix~\cite{li2020dividemix} stage. Additionally, another set of encoders \textit{(Encoder-\{1,2\})} are used to generate the latent image features as depicted in the graphical model from~\cref{fig:graphical_model}. Furthermore, for image reconstruction, the decoders \textit{(Decoder-\{1,2\})} are used by utilizing the continuous Bernoulli loss, and another set of classifiers \textit{(Classifier-\{21,22\})} helps to identify the original noisy labels using the standard cross-entropy loss. 
            %\gustavo{Can we have $\mathbf{y}_1 \sim q_1(Y|X)$ on the left inside the green box, and $\mathbf{y}_2 \sim q_2(Y|X)$ on the right inside the green box?}~\arpit{Updated, added $q^{\rho}$ to make it consistent with the equation}
            %\arpit{added in classifiers-1 block}
            %\gustavo{the classifier-11 and classifier-12, we should have $q(Y|X)$, and not $q(Y|X,\hat{Y})$. We should also index the model, not the variables, so $\mathbf{y}_1 \sim q(Y_1|X,\hat{Y})$ should be $\mathbf{y}_1 = q_1(Y|X)$.  Shouldn't the decoders-2 be called  classifiers-2? For the decoders-1, shouldn't we have $\hat{\mathbf{x}}_1 \sim CB(X;\lambda(z_1,y_1))$? For decoders-2, shouldn't we have $\hat{\mathbf{y}}_1=p_1(\hat{Y}|\hat{\mathbf{x}}_1)$. Please synchronize this figure with the algorithm.}\arpit{updated}
            }            \label{fig:architecture}
\end{figure*}
        
        The objective function in \eqref{eq:minimise_KL_divergence} can then be expanded as:
        \begin{equation}
            \begin{split}
                \mathsf{L}^{(\mathrm{vi})} & = \mathbb{E}_{q(Z | X, Y) q(Y | X)} \left[ - \ln p(X | Z, Y) \right] \\
                & \quad + \mathbb{E}_{q(Y | X)} \left[ - \ln p(\hat{Y} | X, Y) \right]\\
                & \qquad + \mathrm{KL} \left[ q(Y | X) || p(Y) \right] \\
                & \qquad \quad+ \mathbb{E}_{q(Y | X)} \left[ \mathrm{KL} \left[ q(Z | X, Y) || p(Z) \right] \right].
            \end{split}
            \label{eq:objective}
        \end{equation}
        
        \begin{remark}
            The objective function \(\mathsf{L}^{(\mathrm{vi})}\) in \eqref{eq:objective} shares similarity with the loss in variational auto-encoder~\cite{kingma2013auto}. In particular, the first two terms in \eqref{eq:objective} are analogous to the reconstruction loss, while the remaining terms are analogous to the KL loss that regularises the deviation between the posterior \(q\) and its prior.
        \end{remark}
        
        To optimise the objective in \eqref{eq:objective}, both the posteriors \(q(Z | X, Y)\) and \(q(Y | X)\) and priors \(p(Z)\) and \(p(Y)\) must be specified. We assume \(q(Z | X, Y)\) to be a multivariate normal distribution with a diagonal covariance matrix and \(q(Y | X)\) to be a categorical distribution:
        \begin{equation}
            \begin{aligned}[b]
                q(Z | X = \mathbf{x}, \hat{Y} = \hat{\mathbf{y}}) & = \mathcal{N} \left(Z; \bm{\mu}(\mathbf{x}, \hat{\mathbf{y}}), \mathrm{diag}(\bm{\sigma}^{2}(\mathbf{x}, \hat{\mathbf{y}}) \right)\\
                q(Y | X = \mathbf{x}) & = \mathrm{Cat}(Y; \bm{\rho}(\mathbf{x})), 
            \end{aligned}
            \label{eq:probs_encoders}
        \end{equation}
        where the parameters of these distributions are modelled as the outputs of two DNNs. Hereafter, we call the network that models \(q(Y | X)\) the \emph{clean label classifier}, and the  model \(q(Z | X, \hat{Y})\), the \emph{encoder}. 
        
        For the priors, we follow the convention in generative models, especially VAE, to assume \(p(Z)\) as a standard normal distribution, while \(p(Y)\) is a uniform distribution.
        
        Given such assumptions, we can minimise the loss function \(\mathsf{L}^{(\mathrm{vi})}\) in \eqref{eq:objective} w.r.t. the parameters of the two classifiers, the encoder and decoder in \eqref{eq:probs_encoders} and \eqref{eq:probs_decoders}. The obtained clean label classifier that models \(q(Y | X)\) will then be used as the final classifier to evaluate data in the testing set.
        
        \begin{algorithm*}[t]
            \caption{Graphical model approach for learning with label noise}
            \label{algorithm:Proposed_Algo}
            \begin{algorithmic}[1]
                \Procedure{InstanceGM}{$\mathcal{D}, T, \tau $}
                    % long comments to describe input arguments
                    \LComment{\(\mathcal{D}=\{(\mathbf{x}_i, \hat{\mathbf{y}}_i)\}_{i=1}^{|\mathcal{D}|}\): noisy dataset}
                    \LComment{\(T\): total number of epochs}
                    \LComment{\(\tau\): threshold to decide clean or noisy samples used in DivideMix}
                    \State \(q_{1}(Y | X), q_{2}(Y | X) \gets\) \Call{Warmup}{$\mathcal{D}$} \Comment{Warm-up training of 2 clean-label classifiers on noisy dataset}
                    \For{\(e = 1:T\)}
                        \State \(\mathcal{L}_{1}, \mathcal{U}_{1}, \mathcal{L}_{2}, \mathcal{U}_{2} \gets \) \Call{Co-divide}{$ \mathcal{D}, q_{1}(Y | X), q_{2} (Y | X), \tau $}
                        \LComment{Apply Gaussian mixture model on loss values and filter out clean and noisy with a threshold on the likelihood} \LComment{\(\mathcal{L}_{1:2}\) are labelled sets (mostly clean)}
                        \LComment{\(\mathcal{U}_{1:2}\) are unlabelled sets (mostly noisy)}
                        
                        \Statex
                        
                        \State \(\mathsf{L}_{(1)}^{\mathrm{(dm)}} \gets\) \Call{DivideMix Loss}{$ \mathcal{L}_{1}, \mathcal{U}_{1}, q_{1}(Y | X), q_{2}(Y | X) $} \Comment{Calculate training loss in DivideMix}
                        \State \(\mathsf{L}_{(2)}^{\mathrm{(dm)}} \gets\) \Call{DivideMix Loss}{$ \mathcal{L}_{2}, \mathcal{U}_{2}, q_{2}(Y | X), q_{1}(Y | X) $}
                        
                        \For{k = 1:2} \Comment{Calculate loss on each one of the 2 models}
                            \For{each \((\mathbf{x}_{i}, \hat{\mathbf{y}}_{i}) \in \mathcal{L}_{k} \)}
                                \State Compute each instance loss: \(\mathsf{L}_{i}^{\mathrm{(vi)}} \gets\) \Call{Variational-free energy}{$\mathbf{x}_{i}, \hat{\mathbf{y}}_{i}, q_{k}, p_{k}$}
                                \LComment{\(q_{k}\) is the variational posterior}
                                \LComment{\(p_{k}\) denotes prior and data generation}
                            \EndFor
                            \State Compute average loss: \(\mathsf{L}_{(k)}^{(\mathrm{vi})} = \nicefrac{1}{\abs{\mathcal{L}_{k}}} \sum_{i = 1}^{\abs{\mathcal{L}_{k}}} \mathsf{L}_{i}^{\mathrm{(vi)}}\)
                            \State Update model parameters by minimizing \(\mathsf{L}_{(k)} = \mathsf{L}_{(k)}^{(\mathrm{vi})} + \mathsf{L}_{(k)}^{(\mathrm{dm})}\) \Comment{\cref{eq:final_loss_function}}
                        \EndFor
                        % \State Update model parameters by minimizing \(\mathsf{L} = \mathsf{L}_{(1)}^{(\mathrm{vi})} + \mathsf{L}_{(2)}^{(\mathrm{vi})} + \mathsf{L}_{1}^{(\mathrm{dm})} + \mathsf{L}_{2}^{(\mathrm{dm})}\)
                    \EndFor
                    \State \Return \(q_{1}(Y|X)\) \Comment{clean-label classifier}
                \EndProcedure
                
                \Statex
                
                \Function{Variational-free energy}{$\mathbf{x}, \hat{\mathbf{y}}, q, p$} \Comment{Calculate loss in \cref{eq:objective}}
                    \State Sample \(\mathbf{y} \sim q(Y| X = \mathbf{x}_{i})\) \Comment{Sample a clean label from its variational posterior}
                    \State Sample \(\mathbf{z} \sim q(Z | X = \mathbf{x}, \hat{Y} = \hat{\mathbf{y}})\) \Comment{Sample a feature representation from its variational posterior}
                    \State Compute the 1st term in \cref{eq:objective}: \(-\ln p(X = \mathbf{x} | Z = \mathbf{z}, Y = \mathbf{y})\) \Comment{image reconstruction loss}
                    \State Compute the 2nd term in \cref{eq:objective}: \(- \ln p(\hat{Y} = \hat{\mathbf{y}} | X = \mathbf{x}, Y = \mathbf{y})\) \Comment{noisy-label cross-entropy loss}
                    \State Compute the remaining terms in \cref{eq:objective}
                    \State Compute \(\mathsf{L}^{\mathrm{(vi)}}\) as the sum of the above terms as specified in \cref{eq:objective}
                    \State \Return \(\mathsf{L}^{\mathrm{(vi)}}\)
                \EndFunction
            \end{algorithmic}
        \end{algorithm*}

        \begin{remark}
            Optimising the objective function in \eqref{eq:objective} often requires the definition of  hyper-parameters to weight the KL divergences ~\cite{graves2011practical}. However, such weighting mechanism depends on the estimation of the KL divergences weights that is usually achieved with a grid-search using a validation set, making solutions dependent on the dataset. 
            The reason for such weighting mechanism lies at the log-likelhoods used as reconstruction losses. For example, \(-\ln p(X | Z, Y)\) is simply replaced by the corresponding loss functions, such as MSE, without taking the normalisation constants of those likelihood functions into account, resulting in an incorrect balance between reconstruction loss and regularisation. In this paper, we propose the use of the correct form of the log-likelihood, namely the continuous Bernoulli distribution for \(p(X | Z, Y)\) and categorical distribution for \(p(\hat{Y} | X, Y)\), with their normalisation constants. Hence, we no longer need the weighting of the KL divergences, making our proposed method simpler to train.\footnote{More detailed information mentioned in~\cref{sec:motivation_cb}}
            \label{remark:avoid_reweighting}
        \end{remark}
    
    \subsection{Practical implementation}
    \label{sec:practical_implementation}

    In practice, the small loss hypothesis is often used to effectively identify the clean samples in a training set~\cite{han2018co, li2020dividemix}. However, naively implementing such hypothesis using a single model might accumulate error due to sample selection bias. One way to avoid such scenario is to train two models simultaneously where each model is updated using only the clean samples selected by the other model. In this paper, we integrate a similar approach into our modelling presented in \cref{sec:probabilistic_modelling} to solve the label noise problem. 
    In particular, we propose to train two models in parallel, resulting in four classifiers (two for the clean label classifier \(q(Y|X)\) and the other two for noisy labels \(p(\hat{Y}|X,Y)\)), two encoders \(q(Z|X,Y)\) and two decoders \(p(X|Z,Y)\).

    In CausalNL~\cite{yao2021instance}, co-teaching is used as a way to integrate the small loss hypothesis to regularise the clean label classifiers. Co-teaching might, however, limit the capability of the modelling since it only uses samples classified as clean and ignores the other samples classified as noisy. In addition, co-teaching is initially designed for IIN problems, while our focus is on IDN problems. Hence, we propose to integrate DivideMix~\cite{li2020dividemix}, a method based on the small loss hypothesis as shown in \cref{fig:architecture}. This method starts with a warmup stage, and utilizes all training samples after classifying them as clean and noisy (co-divide) using a two-component Gaussian mixture model (GMM). The training samples are used by MixMatch~\cite{berthelot2019mixmatch} -- a semi-supervised classification technique that considers clean samples as labelled and noisy samples as unlabeled.
    DivideMix shows a reasonable efficacy for IDN problems, as shown in~\cref{tab:cifar}.
    % \cref{fig:architecture} shows a diagram of the proposed model.
    
    \begin{table*}[htbp!]
        \caption{Test accuracy (\%) of different methods on CIFAR10 and CIFAR100~\cite{krizhevsky2009learning} under various IDN noise rates. Most results are extracted from~\cite{yao2021instance}, while results with \textsuperscript{*} are reported in their respective papers. Results taken from kMEIDTM~\cite{cheng2022instance} are presented with \textsuperscript{\textdagger}. 
        %\gustavo{I don't see any number in italics} \arpit{changed to \textsuperscript{\textdagger}}
        }
        \label{tab:cifar}
        \centering
        \begin{tabular}{l c c c c c c c c c c}
        \toprule
        \multirow{2}{*}{\bfseries Model} & \multicolumn{5}{c}{\bfseries IDN - CIFAR10} & \multicolumn{5}{c}{\bfseries IDN - CIFAR100} \\ 
        \cmidrule(lr){2-6} \cmidrule(lr){7-11}
        & \textbf{0.20} & \textbf{0.30} & \textbf{0.40} & \textbf{0.45} & \textbf{0.50} & \textbf{0.20} & \textbf{0.30} & \textbf{0.40} & \textbf{0.45} & \textbf{0.50} \\
        \midrule
        CE~\cite{yao2021instance} & 75.81 & 69.15 & 62.45 & 51.72  & 39.42 & 30.42 & 24.15 & 21.45 & 15.23  & 14.42\\
        Mixup~\cite{zhang2017mixup} & 73.17 & 70.02 & 61.56 & 56.45 & 48.95 & 32.92 & 29.76 & 25.92 & 23.13 & 21.31\\
        Forward~\cite{patrini2017making} & 74.64 & 69.75 & 60.21 & 48.81 & 46.27 & 36.38 & 33.17 & 26.75 & 21.93 & 19.27\\
        T-Revision~\cite{xia2019anchor} & 76.15 & 70.36 & 64.09 & 52.42 & 49.02 & 37.24 & 36.54 & 27.23 & 25.53 & 22.54\\
        Reweight~\cite{liu2015classification} & 76.23 & 70.12 & 62.58 & 51.54 & 45.46 & 36.73 & 31.91 & 28.39 & 24.12 & 20.23\\
        PTD-R-V~\cite{xia2020part}\textsuperscript{*}  & 76.58 & 72.77 & 59.50 & \_ & 56.32 & 65.33\textsuperscript{\textdagger} & 64.56\textsuperscript{\textdagger} & 59.73\textsuperscript{\textdagger} & \_ & 56.80\textsuperscript{\textdagger} \\
        Decoupling~\cite{malach2017decoupling} & 78.71 & 75.17 & 61.73 & 58.61 & 50.43 & 36.53 & 30.93 & 27.85 & 23.81 & 19.59\\
        Co-teaching~\cite{han2018co} & 80.96 & 78.56 & 73.41 & 71.60 & 45.92 & 37.96 & 33.43 & 28.04 & 25.60 & 23.97 \\
        MentorNet~\cite{jiang2018mentornet} & 81.03 & 77.22 & 71.83 & 66.18 & 47.89 & 38.91 & 34.23 & 31.89 & 27.53 & 24.15\\
        CausalNL~\cite{yao2021instance} & 81.79 & 80.75 & 77.98 & 79.53 & 78.63 & 41.47 & 40.98 & 34.02 & 33.34 & 32.13\\
        HOC~\cite{zhu2021clusterability}\textsuperscript{*} & 90.03 & \_ & 85.49 & \_ & \_ & 68.82 & \_ & 62.29 & \_ & \_\\
        CAL~\cite{Zhu_2021_CVPR}\textsuperscript{*} & 92.01 & \_ & 84.96 & \_ & \_ & 69.11 & \_ & 63.17 & \_ & \_\\
        kMEIDTM~\cite{cheng2022instance}\textsuperscript{*} & 92.26 & 90.73 & 85.94 & \_ & 73.77 & 69.16 & 66.76 & 63.46 & \_ & 59.18\\
        DivideMix~\cite{li2020dividemix} & 94.80  & 94.60 & 94.53 & 94.08  & 93.04 & 77.07 & 76.33 & 70.80 & 57.78  & 58.61\\
        \midrule
        \rowcolor{Gray!25} \textbf{InstanceGM}  & \textbf{96.68} & \textbf{96.52} & \textbf{96.36} & \textbf{96.15}  & \textbf{95.90} & \textbf{79.69} & \textbf{79.21} & \textbf{78.47} & \textbf{77.49}  & \textbf{77.19}\\
        \bottomrule
        \end{tabular}
        \vspace{-1em}
    \end{table*} 

    \begin{remark}
        Other instance-dependent methods similar to DivideMix~\cite{li2020dividemix}, such as Contrast-to-Divide~\cite{zheltonozhskii2022contrast}, ELR+~\cite{liu2020early}, can also be integrated into our proposed framework. The reason that DivideMix is used is due to its remarkable performance, especially on the IDN setting, and its publicly available implementation.
    \end{remark}

    In general, the loss function for training the proposed model consists of two losses: one is the loss \(\mathsf{L}^{(\mathrm{vi})}\) from the graphical modelling in \eqref{eq:objective}, and the other is the loss to train DivideMix~\cite[Eq. (12)]{li2020dividemix}, denoted as \(\mathsf{L}^{(\mathrm{dm})}\). The whole loss is represented as:
    \begin{equation}
        \mathsf{L} = \mathsf{L}^{(\mathrm{vi})} + \mathsf{L}^{(\mathrm{dm})}
        \label{eq:final_loss_function}
    \end{equation}
    and the training procedure is summarised in \cref{algorithm:Proposed_Algo} and depicted in \cref{fig:architecture}.

    \section{Experiments}
\label{sec:Experimentation}

    In this section, we show the results of extensive experiments on two standard benchmark datasets with IDN, CIFAR10~\cite{krizhevsky2009learning} and CIFAR100~\cite{krizhevsky2009learning} at various noise rates\footnote{Performance degradation at high IDN is presented in~\cref{sec:additional_test}.}, and three real-world datasets, ANIMAL-10N~\cite{song2019selfie}, Red Mini-Imagenet from CNWL~\cite{jiang2020beyond} and CLOTHING-1M~\cite{xiao2015learning}. In \cref{subsec:dataset}, we explain all  datasets mentioned above. In \cref{subsec:implementation}, we discuss all models and their parameters. 
    We compare our approach with state-of-the-art models in IDN benchmarks and real-world datasets
    in \cref{subsec:baseline}.

    \subsection{Datasets}\label{subsec:dataset}
     In both CIFAR10 and CIFAR100, there are \(50k\) training images and \(10k\) testing images with each images of size \(32 \times 32 \times 3\) pixels, where CIFAR10 consists of \(10\) classes, CIFAR100 has  \(100\) classes and both datasets are class-balanced. As CIFAR10 and CIFAR100 datasets do not include label noise by default, we added IDN with noise rates in \(\{0.2, 0.3, 0.4, 0.45,0.5\}\) following the setup proposed by Xia et al.~\cite{xia2020part}.
     
     Red Mini-Imagenet from CNWL~\cite{jiang2020beyond} is a real-world dataset where images and their corresponding labels are crawled from internet at various controllable label noise rates. This dataset is proposed to study real-world noise in controlled settings. In this work, we focus on Red Mini-ImageNet since it shows a realistic type of label noise. Red Mini-ImageNet has \(100\) classes, with each class containing \(600\) images sampled from the ImageNet dataset~\cite{imagenet15russakovsky}. The images are resized to \(32 \times 32\) pixels from the original size of \(84 \times 84\) to have a fair comparison with~\cite{cordeiro2021propmix,xu2021faster}. The noise rates vary from \(0\%\) to \(80\%\), but we use the rates \(20\%\), \(40\%\), \(60\%\) and \(80\%\) to be consistent with the literature~\cite{xu2021faster, yao2021instance, cordeiro2021propmix}.
     
    ANIMAL-10N is another real-world dataset proposed by Song et al.~\cite{song2019selfie}, which contains \(10\) animals with \(5\) pairs having similar appearances (e.g., wolf and coyote, hamster and guinea pig, etc.). The estimated rate of label noise is \(8\%\). There are \(50k\) training images  \(10k\) test images. No data augmentation is used, hence the setup is identical to the one proposed in~\cite{song2019selfie}. 
    
    CLOTHING-1M~\cite{xiao2015learning} is a real-world dataset that comprises \(1\ million\) training apparel images taken from \(14\) categories of online shopping websites. The labels in this dataset are generated from surrounding texts, with an estimated noise of \(38.5\%\). Due to the inconsistency in image sizes, we follow the standard setup in the literature~\cite{han2019deep, li2020dividemix, cordeiro2021propmix} and resize the images to \(256 \times 256\) pixels. This dataset additionally includes \(50k, 14k\), and \(10k\) manually validated clean training, validation, and testing data, respectively. During training, the clean training and validation sets are not used and only the clean testing set is used for assessment.
    
    % WebVision includes $2.4$ million internet-collected images with the same $1000$ classes from ILSVRC12~\cite{deng2009ImageNet} and images reduced to $256 \times 256$ pixels. It offers a clean test set of $50k$ images, with $50$ images per class. We compare our model with the first $50$ classes of the Google image subset, as used in~\cite{li2020dividemix,chen2019understanding}.

    \subsection{Implementation}\label{subsec:implementation}

    %encoders
    All the methods are implemented in PyTorch~\cite{paszke2019pytorch}. For the baseline model DivideMix, all the default hyperparameters are considered as mentioned in original paper by Li et al.~\cite{li2020dividemix}.
    All hyperparameter values mentioned below are from CausalNL~\cite{yao2021instance} and DivideMix~\cite{li2020dividemix} unless otherwise specified.
    The size of the latent representation $Z$ is fixed at \(25\) for CIFAR10, CIFAR100 and Red Mini-Imagenet, \(64\) for ANIMAL-10N, and \(100\) for CLOTHING-1M. 
    For CIFAR10, CIFAR100 and Red Mini-Imagenet, we used non-pretrained PreaAct-ResNet-18 (PRN18)~\cite{he2016identity} as an encoder. 
    VGG-19 is used as an encoder for ANIMAL-10N, following SELFIE~\cite{song2019selfie} and PLC~\cite{zhang2021learning}. For CLOTHING-1M, we used ImageNet-pretrained ResNet-50. \textit{Clean data is not used for training}. 

    %decoders
    % The hyper-parameter $\beta_0$ and $\beta_1$ are fixed to $0.1$, and $\beta_2$ and $\beta_3$ are fixed to \(10^{-5}\) and $0.01$ respectively, as it helps the distribution to be uniform on a small mini-batch. \gustavo{ Which equations use these hyper-parameters?} \arpit{More detailed information are mentioned in~\cref{sec:empirical}.}
    %training
    \begin{table}[t]
        \centering
        \caption{
        Test accuracy (\%) for Red Mini-Imagenet (CNWL)~\cite{jiang2020beyond}. Other model results are as presented in FaMUS~\cite{xu2021faster} and PropMix~\cite{cordeiro2021propmix}. We presented our proposed results with our proposed InstanceGM and with inclusion of self-supervision~\cite{caron2021emerging} in proposed algorithm (InstanceGM-SS). 
        }
        \label{table:RedMini}
        \begin{tabular}{l p{3em} p{3em} p{3em} l}
            \toprule
            \multirow{2}{*}{\bfseries Method} & \multicolumn{4}{c}{\bfseries Noise rate} \\
            \cmidrule{2-5}
            & \bfseries 0.2 & \bfseries 0.4 & \bfseries 0.6 & \bfseries 0.8 \\
            \midrule
            CE~\cite{xu2021faster} & 47.36 & 42.70 & 37.30 & 29.76 \\
            MixUp~\cite{zhang2017mixup} & 49.10 & 46.40 & 40.58 & 33.58 \\
            DivideMix~\cite{li2020dividemix} & 50.96 & 46.72 & 43.14 & 34.50 \\
            MentorMix~\cite{jiang2020beyond} & 51.02 & 47.14 & 43.80 & 33.46 \\
            FaMUS~\cite{xu2021faster} & 51.42 & 48.06 & 45.10 & 35.50\\
            \rowcolor{Gray!25} \textbf{InstanceGM}  & \textbf{58.38} & \textbf{52.24} & \textbf{47.96} & \textbf{39.62} %   \textbf{52.06} \textbf{47.96} old acc
            \\
            \midrule
            \midrule
            \multicolumn{5}{l}{\bfseries With self-supervised learning}\\
            \midrule
            % DivideMix~\cite{} & & & &  \\
            PropMix~\cite{cordeiro2021propmix} & \textbf{61.24} & 56.22 & 52.84 & 43.42\\
            \rowcolor{Gray!25} \textbf{InstanceGM-SS}\tablefootnote{Implementation details are present in~\cref{sec:self_supervision}}  & 60.89  & \textbf{56.37} & \textbf{53.21} & \textbf{44.03} \\ % \textbf{71.31} & \textbf{69.21} & \textbf{66.20} 59.46  for 0.8
            \bottomrule
        \end{tabular}
        % \vspace{-1em}
    \end{table}

    The training of the model used stochastic gradient descent (SGD) for DivideMix stage with momentum of $0.9$, batch size of $64$ and an L2 regularisation whose parameter is $5 \times 10^{-4}$. Additionally, Adam is used to train the VAE part of the model. The training runs for $300$ epochs for CIFAR10, CIFAR100, Red Mini-Imagenet and ANIMAL-10N. The learning rate is $0.02$ which is reduced to $0.002$ at half of the number of training epochs. The WarmUp stage lasts for $10$ epochs for CIFAR10, $30$ for CIFAR100, ANIMAL-10N and Red Mini-Imagenet. For CLOTHING-1M, the WarmUp stage lasts $1$ epoch with batch size of $32$, and training runs for $80$ epochs and learning rate of $0.01$ decayed by a factor of \(10\) after the \(40^{th}\) epoch .

    % \arpit{with decreasing the learning rate by $10$ after epoch $40$.} \gustavo{at same learning rate?}. \cuong{I moved the learning at the end and added the decaying.}
    
    %architectural information -- change the word encoder ans decoder with the probability representation eg: p(x|y,z)
    
    For CIFAR10, CIFAR100~\cite{krizhevsky2009learning}, Red Mini-Imagenet~\cite{jiang2020beyond} and ANIMAL-10N~\cite{song2019selfie}, the encoder has a similar architecture as CausalNL~\cite{yao2021instance}, with $4$ hidden convolutional layers and feature maps containing $32, 64, 128$ and $256$ features. In the decoding stage, we use $4$ hidden layer transposed-convolutional network and the feature maps have $256, 128, 64$ and $32$ features. In Red Mini-Imagenet, we use a similar architecture as CIFAR100 with and without self-supervision~\cite{caron2021emerging}. 
    For CLOTHING-1M~\cite{xiao2015learning}, we use encoder networks with $5$ convolutional layers, and the feature maps contain $32, 64, 128, 256$ and $512$ features. The decoder networks have $5$ transposed-convolutional layers and the feature maps have $512, 256, 128, 64$ and $32$ features.
    
    \subsection{Comparison with Baselines and Measurements}\label{subsec:baseline}

    In this section, we compare our proposed InstanceGM on baseline IDN benchmark datasets in Section~\ref{subsubsec:benchmark}, and we also validate our proposed model on various real-world noisy datasets in Section~\ref{subsubsec:real_world}.
    
    \subsubsection{Instance-Dependent Noise Benchmark Datasets}\label{subsubsec:benchmark}
    
    The comparison between our InstanceGM and recently proposed approaches on CIFAR10 and CIFAR100 IDN benchmarks is shown in Table~\ref{tab:cifar}. 
    Note that the proposed approach achieves considerable improvements in both datasets at various IDN noise rates ranging from $20\%$ to $50\%$. 
    Given that CausalNL represents the main reference for our method, it is important to compare the performance of the two approaches.  For CIFAR10, our method is roughly \(15\%\) better in all noise rates, and for CIFAR100, our method is between \(38\%\) and \(45\%\) better.
    Compared to the current state-of-the-art methods in this benchmark (kMEIDTM~\cite{cheng2022instance} and DivideMix~\cite{li2020dividemix}), our method is around \(2\%\) better in CIFAR10 and between \(2\%\) to almost \(20\%\) better in CIFAR100.
    
    \begin{table}[t]
        \centering
        \caption{Test accuracy (\%) of different methods evaluated on ANIMAL-10N~\cite{song2019selfie} where only noisy data are used to train models. Other models' results are as presented in  Nested-CE~\cite{chen2021boosting}, and results with \textsuperscript{*} are reported in their respective papers}
        \label{table:Animal10N}
        \begin{tabular}{l c}
            \toprule
            \bfseries Method & \bfseries Test Accuracy (\%) \\
            \midrule
            CE~\cite{zhang2021learning} & 79.4 \\
            Nested-Dropout~\cite{chen2021boosting} & 81.3 \\
            CE+Dropout~\cite{chen2021boosting} & 81.3\\
            SELFIE~\cite{song2019selfie}\textsuperscript{*} & 81.8 \\
            PLC~\cite{zhang2021learning}\textsuperscript{*} & 83.4 \\
            Nested-CE~\cite{chen2021boosting} & 84.1 \\
            \midrule
            \rowcolor{Gray!25} \textbf{InstanceGM} & \textbf{84.6} \\
            \bottomrule
        \end{tabular}
        % \vspace{-1em}
    \end{table}

    \subsubsection{Real-world Noisy Datasets}\label{subsubsec:real_world}
    In \cref{table:Animal10N,table:RedMini,table:clothing1M}, we present the results on ANIMAL-10N, Red Mini-Imagenet and CLOTHING-1M, respectively. In general the results show that InstanceGM outperforms or is competitive with the present state-of-the-art models for large-scale web-crawled datasets and small-scale human-annotated noisy datasets.
    %\gustavo{Please try to analyse the results, write in more detail the performance, identify interesting trends, etc.}
    \cref{table:Animal10N} reports the classification accuracy on ANIMAL-10N.
    %on the resized image settings similar to PropMix~\cite{cordeiro2021propmix} \gustavo{this comment is problematic because PropMix doesn't show results on Animal10N, and PropMix isn't on Table~\ref{table:Animal10N}}. 
    We can observe that  InstanceGM achieves slightly better performance than all other baselines. 
    For the other real-world datasets Red Mini-Imagenet and CLOTHING-1M, InstanceGM is competitive, as shown in \cref{table:RedMini,table:clothing1M}, demonstrating its ability to handle real-world IDN problems. In particular, ~\cref{table:RedMini} shows the results on Red Mini-Imagenet using two set-ups: 1) without pre-training (top part of the table), and 2) with self-supervised (SS) pre-training (bottom part of the table).
    The SS pre-training is based on DINO~\cite{caron2021emerging} with the unlabelled Red Mini-Imagenet dataset, allowing a fair comparison with PropMix~\cite{cordeiro2021propmix}, which uses a similar SS pre-training.
    %with InstanceGM and for the fair comparison with PropMix~\cite{cordeiro2021propmix}, which used self-supervision, we provided results of InstanceGM-SS. InstanceGM-SS is the InstsanceGM model with self-supervision DINO~\cite{caron2021emerging}.
    Without SS pre-training, our InstanceGM is substantially superior to recently proposed approaches.
    With SS pre-training, results show that InstanceGM can improve its performance, allowing us to achieve state-of-the-art results on Red Mini-Imagenet.

    \begin{table}
        \centering
        \caption{Test accuracy (\%) for  competing methods on CLOTHING-1M~\cite{xiao2015learning}. The accuracy of the baseline models (CausalNL and DivideMix)  are in italics. 
        Results of other models are from their respective papers. In the experiments only noisy labels are use for training. Top results with \(1\%\) accuracy are highlighted in bold.}
        \label{table:clothing1M}
        \begin{tabular}{l c}
            \toprule
            \bfseries Method & \bfseries Test Accuracy (\%) \\
            \midrule
            % Co-teaching~\cite{han2018co} & 60.15 \\
            % CE~\cite{yao2021instance} & 68.88 \\
            % T-Revision~\cite{xia2019anchor} & 70.97 \\
            CausalNL~\cite{yao2021instance} & \textit{72.24} \\
            IF-F-V~\cite{jiang2021information} & 72.29\\
            % ELR~\cite{liu2020early} & 72.87 \\
            % PropMix~\cite{cordeiro2021propmix} & 74.30\\
            % C2D~\cite{zheltonozhskii2022contrast} & 74.30 \\
            % FaMUS*~\cite{xu2021faster} & 74.40 \\
            \textbf{DivideMix}~\cite{li2020dividemix} & \textit{\textbf{74.76}} \\
            % ELR+*~\cite{liu2020early} & 74.81\\
            % RRL*~\cite{li2021learning} & 74.84 \\
            \textbf{Nested-CoTeaching}~\cite{chen2021boosting} & \textbf{74.90}\\
            % AugDesc*~\cite{nishi2021augmentation} & 75.11\\
            \midrule
            \rowcolor{Gray!25} \textbf{InstanceGM}  & \textbf{74.40} \\
            \bottomrule
        \end{tabular}
        % \vspace{-1em}
    \end{table}

    % \setlength{\tabcolsep}{4pt}
    %%
    % \begin{table}
    %     \centering
    %     \caption{Test accuracy (\%) for the competing methods on Webvision. In the experiments only noisy labels are exploited to train for $100$ epochs. Baselines comes from~\cite{li2020dividemix}. Top methods within $1\%$ are in bold.}
    %     \label{table:WebVision}
    %     \begin{tabular}{l p{4em} p{4em}}
    %         \toprule
    %         \bfseries Method & \bfseries Top-1 & \bfseries Top-5 \\
    %         \midrule
    %         Co-teaching~\cite{} & 63.58 & 85.20 \\
    %         ELR+~\cite{} & 77.78 & 91.68 \\
    %         DivideMix~\cite{li2020dividemix} & \textbf{77.32} & \textbf{91.64} \\
    %         \rowcolor{Gray!25} \textbf{Ours} & \_ & \_ \\
    %         \bottomrule
    %     \end{tabular}
    % \end{table}

    \section{Ablation Study}
    
    We show the ablation study of our proposed method on CIFAR10~\cite{krizhevsky2009learning}, under IDN noise rate of \(0.5\) and ANIMAL-10N~\cite{song2019selfie}. On Table~\ref{tab:ablation_cifar}, the performance of CausalNL~\cite{yao2021instance} is relatively low, which can be explained by the small number of clean samples used by co-teaching~\cite{han2018co}, and  the use of MSE for image reconstruction loss\footnote{line 80 in https://github.com/a5507203/IDLN/blob/main/causalNL.py}.
    % \gustavo{please confirm that they use MSE error}\arpit{yes, code checked https://github.com/a5507203/IDLN/blob/main/causalNL.py line number 80}.
    We argue that replacing co-teaching~\cite{han2018co} by DivideMix~\cite{li2020dividemix} will improve classification accuracy because it allows the use of the whole training set. 
    To demonstrate that, we take CausalNL~\cite{yao2021instance} and replace its co-teaching by DivideMix, but keep the MSE reconstruction loss -- this model is named CausalNL + DivideMix (w/o continuous Bernoulli).
    Note that this allows a $\approx 10\%$ accuracy improvement from CausalNL, but the use of MSE reconstruction loss can still limit classification accuracy.
    Hence, by replacing the MSE loss by the continuous Bernoulli loss for image reconstruction, we notice a further $\approx 7\%$ accuracy improvement.  
    %, which is not helpful \arpit{due to the usage of mean square error while modelling. These types of modelling cause pervasive error~\cite{loaiza2019continuous} due to the variation in handling of image pixels. (Remark~\ref{remark})} \gustavo{please repeat the reason here}.
    
    \begin{table}[t]
        \centering
        \caption{This ablation study shows the test accuracy \(\%\) on CIFAR10 under IDN at noise rate \(0.5\). First, we show the result of CausalNL~\cite{yao2021instance}. Second, we show the result of CausalNL~\cite{yao2021instance} with Co-teaching~\cite{han2018co} replaced by DivideMix~\cite{li2020dividemix} (without Continuous Bernoulli reconstruction). Then at last we show the results of our proposed algorithm InstanceGM. 
        %\gustavo{This table can be single column.}\arpit{done} 
        }
        \label{tab:ablation_cifar}
        \begin{tabular}{l c}
            \toprule
            \bfseries Method &  \bfseries Test Accuracy (\%) \\
            \midrule
            % CausalNL~\cite{yao2021instance} + continuous Bernoulli reconstruction  & 66.90\\
            CausalNL~\cite{yao2021instance} & 78.63 \\
            CausalNL~\cite{yao2021instance} + DivideMix~\cite{li2020dividemix}  & \multirow{2}{*}{88.62} \\
            \quad * w/o continuous Bernoulli   & \\
            \rowcolor{Gray!25} \textbf{InstanceGM} & \textbf{95.90} \\
            \bottomrule
        \end{tabular}
    \end{table}
    
    For ANIMAL-10N~\cite{song2019selfie}, we test InstanceGM with various backbone networks (VGG~\cite{simonyan2014very}, ResNet~\cite{he2016deep}, and ConvNeXt~\cite{liu2022convnet}) and the results are displayed in~\cref{tab:ablation_animal}. 
    %For the comparison between various architectures, we tried VGG~\cite{simonyan2014very}, ResNet~\cite{he2016deep}, ConvNeXt~\cite{liu2022convnet}. 
    Due to the architectural differences, ConvNeXt~\cite{liu2022convnet} performed best on our proposed algorithm, but for a fair comparison with the other models, we use the VGG backbone~\cite{simonyan2014very} results in~\cref{table:Animal10N}.

    \begin{table}[t]
        \centering
        \caption{This ablation study shows the test accuracy \(\%\) on ANIMAL-10N using various architectures (without self-supervision), including ResNet~\cite{he2016deep}, VGG~\cite{simonyan2014very} and ConvNeXt~\cite{liu2022convnet} with InstanceGM. \cref{table:Animal10N}, reported the results of VGG~\cite{simonyan2014very} to provide a fair comparison with other methods.}
        \label{tab:ablation_animal}
        \begin{tabular}{l  c}
            \toprule
            \bfseries Method & \bfseries Test Accuracy (\%) \\
            \midrule
            InstanceGM with ResNet~\cite{he2016deep} & 82.2 \\
            InstanceGM with VGG~\cite{simonyan2014very} & 84.6\\
            InstanceGM with ConvNeXt~\cite{liu2022convnet} & 84.7\\
            \bottomrule
        \end{tabular}
        % \vspace{-1em}
    \end{table}

    \section{Conclusion}
\label{sec:conclusion}

% need to add discussion here

In this paper, we presented an instance-dependent noisy label learning algorithm method, called InstanceGM. 
%We used two encoders and two decoders for the generative approach, considering the fact that instances are useful for inferring the correct labels. 
InstanceGM explores generative and discriminative models~\cite{yao2021instance}, where for the generative model, we replace the usual MSE image reconstruction loss by the continuous Bernoulli reconstruction loss~\cite{loaiza2019continuous} that improves the training process, and for the discriminative model, we replace co-teaching by DivideMix~\cite{li2020dividemix} to enable the use of clean and noisy samples during training.
We performed extensive experiments on various IDN benchmarks, and our results on CIFAR10, CIFAR100, Red Mini-Imagenet, ANIMAL-10N outperform the results of state-of-the-art methods, particularly in high noise rates and are competitive for CLOTHING-1M. 
The ablation study clearly shows the importance of the new continuous Bernoulli reconstruction loss~\cite{loaiza2019continuous} and DivideMix~\cite{li2020dividemix}, with both improving classification accuracy from CausalNL~\cite{yao2021instance}.
% In our future work, we can extend to incorporation of other types of discriminative classifiers. 
% Moreover, we will also study the theoretical properties of the data generation process and the assumptions made in the proposed algorithm
% \gustavo{please be more specific regarding these two points.}\arpit{removed the discussion from section and made it purely conclusion}  

    {\small
    \bibliographystyle{ieee_fullname}
    \bibliography{references}
    }
    
    \onecolumn
\appendix
\section*{Appendix}
The appendix is organised as follows:
\begin{itemize}
    % \item \cref{sec:empirical} includes the empirical explanation of variational inference.
    % \item~\cref{sec:detailed_experimental_setup} shows additional experimental details in detailed view.
    \item \cref{sec:self_supervision} presents a detailed description of InstanceGM-SS and experimental details of self-supervision. \cref{subsec:exp_self_dino} contains the experimental details for the self-supervision method DINO, and \cref{subsec:exp_self_instances} contains the experimental details of InstanceGM with self-supervision (InstaceGM-SS).
    \item \cref{sec:motivation_cb} shows the motivation behind the use of the continuous Bernoulli distribution.
    % \item \cref{sec:training_algorithm_dividemix} includes the co-divide algorithm used by our proposed work.
    \item \cref{sec:additional_test} shows the results on high IDN rates for CIFAR10 dataset. 
    % Moreover, \cref{sec:sample_test} contains samples of results on ANIMAL-10N (\cref{fig:chimp}) and on CIFAR100 (\cref{fig:cifar}).
    
\end{itemize}

\section{Self-supervision and experimental details}\label{sec:self_supervision}
    % For the Red mini-Imagenet, we adapted the self-supervision for classifiers in InstanceGM and denoted it with InstanceGM-SS in Table~\ref{table:RedMini}. This shows the easy adaptability of the proposed algorithm and fair comparison with the model using self-supervision stage like PropMix~\cite{cordeiro2021propmix}. We used the DINO~\cite{caron2021emerging} for the self-supervision.
    
    In addition to training the proposed method from scratch, we also adapt self-supervised learning to pre-train the feature extractor part in the classifier \(q(Y | X)\), denoted as InstanceGM-SS in \cref{table:RedMini}. In particular, we employ DINO~\cite{caron2021emerging} to self-supervisedly learn a feature extractor using the unlabelled data from the training set of Red Mini-Imagenet (DINO is a self-supervision method that uses self-distillation). Such integration allows our proposed method to be fairly compared with other label noise learning approaches that rely on self-supervision, such as PropMix~\cite{cordeiro2021propmix}.

    \subsection{Experimental details of self-supervision DINO}\label{subsec:exp_self_dino}
        We trained the  self-supervised model on Red Mini-Imagenet for \(500\) epochs on  PreAct-ResNet-18 (PRN18). We used the same set of hyper-parameters as provided by DINO. The method follows the teacher-student setting where the weights of the teacher network are exponentially weighted averaged from the student network~\cite{he2020momentum}. It includes the teacher model temperature for warmup as \(0.04\) and \(0.07\) for training, and warmup teacher epochs as \(50\). The L2 weight decay regularisation is \(0.000001\), and batch size is \(51\) per gpu. The initial learning rate is set to \(0.3\) and the minimum learning rate is set to \(0.0048\), with the training warm up number of epochs set as \(10\). In addition,DINO needs various augmented views of the input image. It includes multi-crop strategy~\cite{caron2020unsupervised} with high-resolution global and low-resolution local views.
        The two versions of global crops views are considered with scale values of \(0.14\) and \(1\). Moreover, the six different local crops views are considered having scale values of \(0.05\) and \(0.14\), with teacher momentum as \(0.996\). %\gustavo{also unclear}

    \subsection{Experimental details of InstanceGM-SS}\label{subsec:exp_self_instances}
        When we use the self-supervised trained classifier for InstanceGM-SS, we slightly change  the settings to train the Red Mini-Imagenet, and results could be find in~\cref{table:RedMini}. 
        %We used the settings similar to the ones for training Red Mini-ImageNet without self-supervision. 
        In particular, we use the self-supervised  PreAct-ResNet-18 as a classifier with the latent representation Z of size \(25\). We train the network for \(80\) epochs with the learning rate reduced by 10 after \(50\) epochs. The warmup stage is reduced to \(15\) epochs. Otherwise, the previous settings for Red Mini-ImageNet without self-supervision are kept the same.

\section{Motivation of using continuous Bernoulli distribution}\label{sec:motivation_cb}
    To explain the motivation behind the use of the continuous Bernoulli likelihood for image reconstruction, we refer to the variational inference technique. In particular, we denote \(x\) as an observable variable, e.g., input images, while \(z\) as a hidden (or latent) variable. For simplicity, we assume that both \(x\) and \(z\) are scalars. In variational inference, e.g. VAE, the objective is to maximise the evidence lower bound (ELBO) or minimise the variational-free energy w.r.t. \(\phi\) -- the parameter of the variational posterior \(q_{\phi}(z | x)\):
    \begin{equation}\label{eq:vae_reconst}
        \min _{\phi} \underbrace{\mathbb{E}_{q_{\phi}(z | x)}[-\ln p_{\theta}(x | z)]}_{\text {\textcolor{BrickRed}{Reconstruction loss} }} + \beta \, \mathrm{KL}[q_{\phi}(z | x) || p(z)],
    \end{equation}
    where \(p(z)\) is the prior of \(z\) (for example, a standard Gaussian distribution $\mathcal{N}(0, \mathbf{I})$),  \(\beta \in \mathbb{R}_{+}\) is a re-weighting factor. In theory, \(\beta = 1\), and we use \(\beta\) here to explain common practice in VAE which is described in the following.

    The second term in \eqref{eq:vae_reconst} could be evaluated with a closed-form formula for some simple cases of \(q_{\phi}(z | x)\) and \(p(z)\) or approximated using Monte Carlo sampling. Thus, we would focus on the explanation of the first term -- often known as reconstruction loss. Depending on how \(p_\theta(x | z)\) is modelled, we could have different reconstruction losses, as explained below.

    \subsection{Gaussian likelihood}
        If \(p_{\theta}(x | z)\) is a Gaussian distribution: \(p_{\theta}(x | z) = \mathcal{N}(x; \mu(z), \sigma^{2}(z)))\), the negative log-likelihood term in \eqref{eq:vae_reconst} can be written as:
        \begin{equation}\label{eq:mse_full}
            -\ln p_{\theta}(x | z) = \ln (\sigma(z) \sqrt{2\pi}) + \frac{1}{\sigma^{2}(z)}\textcolor{PineGreen}{(\mathbf{x - \mu(z))^{2}}}.
        \end{equation}
        The correct form of the reconstruction loss in \eqref{eq:mse_full} contains two terms including a \say{weighted} MSE term. However, common practice simply replaces the whole \(-\ln p_{\theta}(x | z)\) by \(\textcolor{PineGreen}{(\mathbf{x - \mu(z))^{2}}}\), resulting in an incorrect formula. As a result, it requires to fine-tune \(\beta\) to some small value to balance the contributions of the first and second terms in \eqref{eq:vae_reconst}.

    \subsection{Bernoulli likelihood}
        If \(p_{\theta}(x | z)\) is a Bernoulli distribution: $p_{\theta}(x | z) = \mathcal{B}(\lambda_{\theta}(z))$ where $x \in {\{0,1\}}$ and \(\lambda_{\theta}(z) \in [0, 1]\), then the negative log-likelihood in \eqref{eq:vae_reconst} is: 
        \begin{equation}\label{eq:bernoulli_vae}
            - \ln p_{\theta}\left(x | z\right) = -x \ln \lambda_{\theta}\left(z\right) - \left(1 - x\right) \ln \left(1 - \lambda_{\theta}\left(z\right)\right),
        \end{equation}
        resulting in the binary cross-entropy loss (BCE)~\cite{creswell2017denoising}.
        
        Simply implementing the reconstruction loss as BCE results in the pervasive error since the input \(x\) must be in \{0, 1\}~\cite{loaiza2019continuous}, which is applicable for black and white images only.

        % In~\cref{eq:bernoulli_vae}  $\lambda_{\theta}$ is $\theta$ parameterized neural network. Data considered is binary with the likelihood conditioned $p_{\theta}(x_{n}|z_{n})$ is selected to $\mathcal{B}(\lambda_{\theta}(z_{n}))$. Here, $\mathcal{B}(\lambda_{\theta})$ is the product of non-dependent Bernoulli distributions $(\lambda_{\theta} \in [0,1])$. As a result for colored dataset, it leads to an imbalance between the \say{implemented} reconstruction loss and the KL divergence, requiring re-weighting through \(\beta\). Such ad-hoc approach, however, adds complexity to the training, making the reconstruction loss more error prone~\cite{higgins2016beta}. It fails to evaluate the probabilistic inference, as it is lacks the normalised distribution.
    
    \subsection{Continuous Bernoulli likelihood}
        For colour images, although one can model \(p_{\theta}(x | z)\) as a Gaussian distribution shown in \eqref{eq:mse_full}, it might be a suboptimal choice since the support of the Gaussian distribution is un-bounded, while image data is bounded. Thus, we use the continuous Bernoulli distribution to model \(p_{\theta}(x | z)\)~\cite{loaiza2019continuous} since the continuous Bernoulli distribution is supported in \([0, 1]\) with only one parameter:
        % \begin{equation}\label{eq:continuous_bernoulli}
        %     x \sim \mathcal{C B}(\lambda_{\theta}) \Longleftrightarrow p_{\theta}(x_{n} \mid \lambda_{\theta}) \propto \tilde{p}_{\theta}(x_{n} \mid \lambda_{\theta})=\lambda_{\theta}^{x_{n}}(1-\lambda_{\theta})^{1-x_{n}}
        % \end{equation}
        % We denote the Bernoulli distribution as $\tilde{p}_{\theta}(x_{n}|\lambda_{\theta}) = \lambda_{\theta}^{x_{n}}(1-\lambda_{\theta})^{(1-x_{n})}$ similar to Loaiza et al.~\cite{loaiza2019continuous}. For $\lambda_{\theta} \in (0,1)$, the continuous Bernoulli is 
        % $0<\int_{0}^{1} \tilde{p}_{\theta}(x_{n} \mid \lambda_{\theta}) d x_{n}<\infty$. Additionally, the density function for $X_{n} \sim \mathcal{C B}(\lambda_{\theta})$, and the expected values are transformed into:
        \begin{equation}
        \begin{aligned}[b]
        p_{\theta}(x | \lambda_{\theta}) &=\textcolor{RawSienna}{C(\lambda_{\theta})} \lambda_{\theta}^{x} (1 - \lambda_{\theta})^{1 - x}, \text { where } \textcolor{RawSienna}{C(\lambda_{\theta})}= \begin{cases}\frac{2 \tanh ^{-1}(1-2 \lambda_{\theta})}{1-2 \lambda_{\theta}} & \text { if } \lambda_{\theta} \neq 0.5 \\
        2 & \text { otherwise.}\end{cases} \\
        % \mu(\lambda_{\theta}) &:=E[X_{n}]= \begin{cases}\frac{\lambda_{\theta}}{2 \lambda_{\theta}-1}+\frac{1}{2 \tanh ^{-1}(1-2 \lambda_{\theta})} & \text { if } \lambda_{\theta} \neq 0.5 \\
        % 0.5 & \text { otherwise }\end{cases}
        \end{aligned}
        \end{equation}
        
        Note that one could also use the Beta distribution whose support space is also \([0, 1]\). The advantage of using the continuous Bernoulli distribution is the simplicity since we need only one parameter per pixel, while the Beta distribution requires double the number of parameters.

\section{Experimental Results on CIFAR10 at High IDN Levels}\label{sec:additional_test}
% \subsection{Experimental Results on CIFAR10 at High IDN Levels}\label{sec:high_exp}
We investigated the performance of InstanceGM on high IDN levels including \(0.7, 0.8\) and \(0.9\). We provided the test classification accuracy on CIFAR10 on~\cref{table:high_cifar}. The competing model results are from~\cite{jiang2021information}. Our InstanceGM shows superior results even in such high noise rate problems. Note that all results at these high noise rate problems are not good but the performance degradation for InstanceGM is lower, compared to the other models.

\begin{table}[h]
        \centering
        \caption{
        Test accuracy (\%) for CIFAR10 at high IDN rates. All the mentioned results of other methods are as presented in the paper~\cite{jiang2021information}. 
        %We presented our proposed results with our proposed InstanceGM.
        }
        \label{table:high_cifar}
        \begin{tabular}{l p{3em} p{3em} p{3em} l}
            \toprule
            \multirow{2}{*}{\bfseries Method} & \multicolumn{3}{c}{\bfseries IDN - CIFAR10} \\
            \cmidrule{2-4}
            & \bfseries 0.7 & \bfseries 0.8 & \bfseries 0.9 \\
            \midrule
            % PTD-F~\cite{xia2020part} & 20.25 & 13.48 & 09.22\\
            PTD-F-V~\cite{xia2020part} & 20.35 & 13.58 & 09.44\\
            % PTM-F~\cite{jiang2021information} & 16.79 & 14.13 & 10.34\\
            PTM-F-V~\cite{jiang2021information} & 18.95 & 13.89 & 10.57\\
            % IF-F~\cite{jiang2021information} & 21.26 & 15.78 & 10.58\\
            IF-F-V~\cite{jiang2021information} & 21.09 & 16.72 & 10.86\\
            DivideMix~\cite{li2020dividemix} & 22.13 & 08.10 & 04.08\\
            \midrule
            \rowcolor{Gray!25} \textbf{InstanceGM}  & \textbf{47.23} & \textbf{29.30} & \textbf{11.01}\\ %   \textbf{52.06} \textbf{47.96} old acc
            \bottomrule
        \end{tabular}
    \end{table}

\end{document}